\documentclass[10pt,twocolumn,letterpaper]{article}

\usepackage{iccv}
\usepackage{times}
\usepackage{epsfig}
\usepackage{graphicx}
\usepackage{amsmath}
\usepackage{amssymb}

\usepackage{amsmath,amsfonts,bm}

\def\eqref#1{equation~\ref{#1}}

\def\1{\bm{1}}

\DeclareMathAlphabet{\mathsfit}{\encodingdefault}{\sfdefault}{m}{sl}
\SetMathAlphabet{\mathsfit}{bold}{\encodingdefault}{\sfdefault}{bx}{n}

\usepackage{pgfplots}
\usepackage{pgfplotstable}
\usepackage{amssymb}
\usepackage{wrapfig}
\usepackage{algorithm}
\usepackage{listings}
\usepackage{subfigure}
\usepackage{multirow}
\usepackage{makecell}

\usepackage[accsupp]{axessibility}  

\usepackage{pifont}
\usepackage[pagebackref=true,breaklinks=true,letterpaper=true,colorlinks,bookmarks=false]{hyperref}

\pgfplotsset{compat=1.8}
\usetikzlibrary{pgfplots.groupplots}
\definecolor{mycitecolor}{rgb}{0, 0.4, 0.7}
\definecolor{chromeyellow}{rgb}{1.0, 0.65, 0.0}
\definecolor{darkcerulean}{rgb}{0.03, 0.27, 0.49}
\definecolor{darkorange}{rgb}{1.0, 0.55, 0.0}
\definecolor{darkmidnightblue}{rgb}{0.0, 0.2, 0.4}
\definecolor{internationalorange}{rgb}{1.0, 0.31, 0.0}
\definecolor{internationalkleinblue}{rgb}{0.0, 0.18, 0.65}
\definecolor{lightsalmon}{rgb}{1.0, 0.63, 0.48}
\definecolor{mangotango}{rgb}{1.0, 0.51, 0.26}
\definecolor{mayablue}{rgb}{0.45, 0.76, 0.98}
\definecolor{majorelleblue}{rgb}{0.38, 0.31, 0.86}
\definecolor{mediumelectricblue}{rgb}{0.01, 0.31, 0.59}
\definecolor{bananamania}{rgb}{0.98, 0.91, 0.71}
\definecolor{mossgreen}{rgb}{0.68, 0.87, 0.68}
\definecolor{mossgreen2}{rgb}{0.58, 0.80, 0.48}
\definecolor{mintgreen}{rgb}{0.6, 1.0, 0.6}
\definecolor{palegreen}{rgb}{0.6, 0.98, 0.6}
\definecolor{bananayellow}{rgb}{1.0, 0.88, 0.21}
\definecolor{bluebell}{rgb}{0.64, 0.64, 0.82}
\definecolor{carnationpink}{rgb}{1.0, 0.65, 0.79}
\definecolor{babyblue}{rgb}{0.54, 0.81, 0.94}

\definecolor{mycolor}{rgb}{0, 0.4, 0.7}
\newlength\savewidth\newcommand\shline{\noalign{\global\savewidth\arrayrulewidth
  \global\arrayrulewidth 1pt}\hline\noalign{\global\arrayrulewidth\savewidth}}
\newcommand{\tablestyle}[2]{\setlength{\tabcolsep}{#1}\renewcommand{\arraystretch}{#2}\centering}
\newcolumntype{x}[1]{>{\centering\arraybackslash}p{#1pt}}
\newcolumntype{y}[1]{>{\raggedright\arraybackslash}p{#1pt}}
\newcolumntype{z}[1]{>{\raggedleft\arraybackslash}p{#1pt}}
\newcommand{\reshl}[3]{
\tablestyle{1.2pt}{1}
\begin{tabular}{z{20}y{25}}
{#1} & \fontsize{7.5pt}{1em}\selectfont{\textcolor{mycolor}{(${#2}$\textbf{#3})}} 
\end{tabular}}
\renewcommand{\paragraph}[1]{\vspace{1.25mm}\noindent\textbf{#1}}
\iccvfinalcopy 

\def\iccvPaperID{2612} 

\ificcvfinal\fi

\begin{document}

\title{Group DETR: Fast DETR Training with Group-Wise One-to-Many Assignment}
\author{Qiang Chen$^{1*}$, Xiaokang Chen$^{2}$\thanks{Equal contribution.}~~, Jian Wang$^{1}$, Shan Zhang$^{3}$ \\
Kun Yao$^{1}$, Haocheng Feng$^{1}$, Junyu Han$^{1}$, Errui Ding$^{1}$, Gang Zeng$^{2}$, Jingdong Wang$^{1}$\thanks{Corresponding author.} \\
$^{1}$Baidu VIS \\
$^{2}$Key Lab. of Machine Perception (MoE), School of IST, Peking University \\
$^{3}$Australian National University \\
\texttt{\{chenqiang13,wangjian33\}}@baidu.com \\
\texttt{\{fenghaocheng,hanjunyu,dingerrui,wangjingdong\}}@baidu.com \\
\texttt{\{pkucxk,gang.zeng\}}@pku.edu.cn, \texttt{shan.zhang}@anu.edu.au
}

\maketitle
\ificcvfinal\thispagestyle{empty}\fi

\begin{abstract}

Detection transformer (DETR) relies on one-to-one assignment, assigning one ground-truth object to one prediction, for end-to-end detection without NMS post-processing. It is known that one-to-many assignment, assigning one ground-truth object to multiple predictions, succeeds in detection methods such as Faster R-CNN and FCOS. While the naive one-to-many assignment does not work for DETR, and it remains challenging to apply one-to-many assignment for DETR training. In this paper, we introduce Group DETR, a simple yet efficient DETR training approach that introduces a group-wise way for one-to-many assignment. This approach involves using multiple groups of object queries, conducting one-to-one assignment within each group, and performing decoder self-attention separately. It resembles data augmentation with automatically-learned object query augmentation. It is also equivalent to simultaneously training parameter-sharing networks of the same architecture, introducing more supervision and thus improving DETR training. The inference process is the same as DETR trained normally and only needs one group of queries without any architecture modification. Group DETR is versatile and is applicable to various DETR variants. The experiments show that Group DETR significantly speeds up the training convergence and improves the performance of various DETR-based models. Code will be available at \url{https://github.com/Atten4Vis/GroupDETR}.
\vspace{-3mm}
\end{abstract}

\section{Introduction}
Detection Transformer (DETR)~\cite{detr} conducts end-to-end object detection without the need for many hand-crafted components, such as non-maximum suppression (NMS)~\cite{hosang2017learning} and anchor generation~\cite{ren2015faster,lin2017focal,redmon2018yolov3}. The architecture consists of a CNN~\cite{he2016deep} and transformer encoder~\cite{vaswani2017attention}, and a transformer decoder that consists of self-attention, cross-attention and FFNs, followed by class and box prediction FFNs. During training, one-to-one assignment, where one ground-truth object is assigned to one single prediction, is applied for learning to only promote the predictions assigned to ground-truth objects, and demote the duplicate predictions.
\begin{figure}
\centering
\includegraphics[width=.49\textwidth]{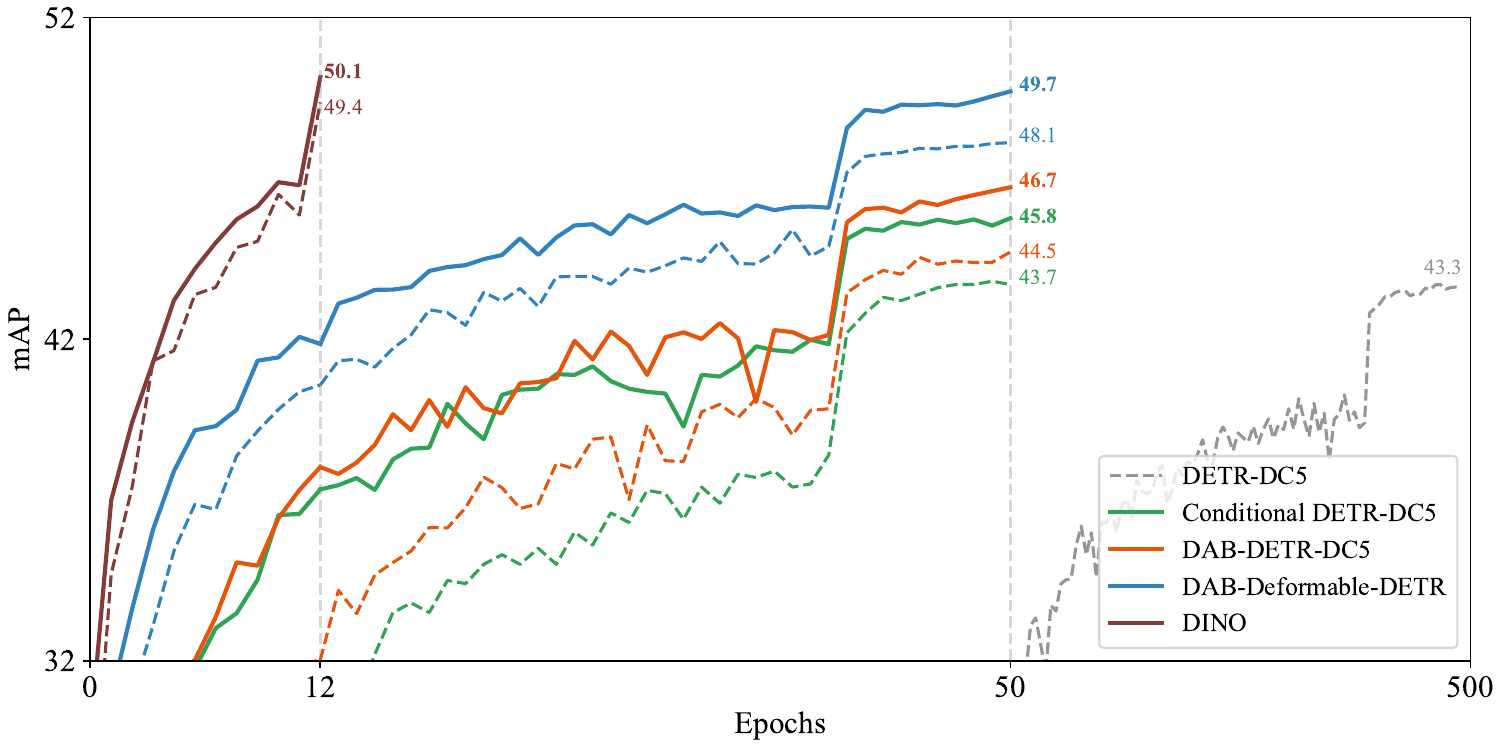}
\caption{\textbf{Group DETR accelerates the training process for DETR variants.} 
The training convergence curves are obtained 
on COCO {\em val2017}~\cite{lin2014coco} 
with ResNet-$50$~\cite{he2016deep}. 
Dashed and bold curves correspond
to the baseline models and the Group DETR counterparts. Best viewed in color.}
\label{fig:curves}
\vspace{-3mm}
\end{figure}

This work explores the solutions to accelerate the DETR training process. Previous solutions contain two main lines. The one line is to modify~\emph{cross-attention} so that informative image regions are selected for effectively and efficiently collecting the information from image features. Example methods include sparse sampling, through deformable attention~\cite{deformable-detr}, and spatial modulations with modifying object queries~\cite{gao2021fast,meng2021conditional,chen2022conditional,wang2021anchor,yao2021efficient,liu2022dab,gao2022adamixer}. The other line is to stabilize \emph{one-to-one assignment} during training, e.g., feeding ground-truth bounding boxes with noises into transformer decoder~\cite{li2022dn,zhang2022dino}.

We are interested in the second line. Instead of focusing on stabilizing the assignment like DN-DETR~\cite{li2022dn}, we study the assignment scheme for efficient DETR training from a new perspective: introducing more supervision. It has been proven that assigning one ground-truth object to multiple predictions, i.e., one-to-many assignment, is successful in traditional object detection methods, e.g., Faster R-CNN~\cite{ren2015faster} and FCOS~\cite{tian2019fcos} with more anchors and pixels assigned to one ground-truth object. Unfortunately, naive one-to-many assignment does not work for DETR training. It remains a challenge to apply one-to-many assignment to DETR training.
\begin{figure*}[h]
\centering
\footnotesize
\subfigure[]{
\includegraphics[width=0.315\textwidth]{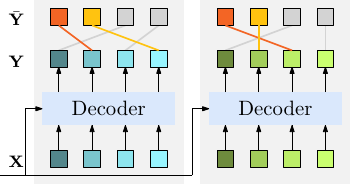}}~~~~~
\subfigure[]{\includegraphics[width=0.315\textwidth]{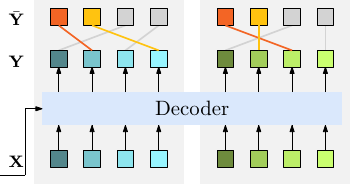}}~~~~~
\subfigure[]{\includegraphics[width=0.315\textwidth]{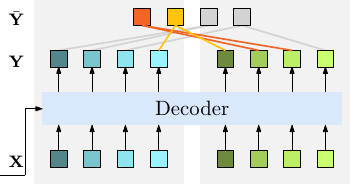}}
\caption{\textbf{Architecture illustration.} 
(a) Our Group DETR: group-wise one-to-many assignment and separate self-attention, architecturally equivalent to parallel decoder. (b) Group-wise one-to-many assignment only. (c) Naive one-to-many assignment. We use two groups of $4$ object queries as an example. $\mathbf{X}$: image features; $\mathbf{Y}$: predictions; $\bar{\mathbf{Y}}$: ground-truth objects, where two color boxes mean two objects and two gray boxes mean dummy objects (no objects). The color lines between $\mathbf{Y}$ and $\bar{\mathbf{Y}}$ correspond to the assignment for ground-truth objects, and the gray lines for dummy objects. For clarity, the predictors are not explicitly included.}
\label{fig:groupdetrandothers} 
\end{figure*}

We present a simple yet efficient DETR training approach that uses a group-wise way for one-to-many assignment, called Group DETR. Our approach is based on that end-to-end detection with successful removal of NMS post-processing for DETR comes from the joint effect of two components~\cite{detr,meng2021conditional}: decoder self-attention, which collects the information of other predictions, and one-to-one assignment, which expects to learn to score one prediction higher and other duplicate predictions lower for one ground-truth object.

Our approach adopts $K$ groups of object queries, and introduces~\emph{group-wise one-to-many assignment}. This assignment scheme conducts one-to-one assignment within each group of object queries, resulting in that one ground-truth object is assigned to multiple predictions. It is  encouraged that the prediction assigned to the ground-truth object gets a high score, and other duplicate predictions from the same group of queries get low scores. In other words, the predictions make competition within each group. Thus, our approach uses~\emph{separate self-attention}, i.e., self-attention is done for each group separately, eliminating the influence of predictions from other groups and easing DETR training. Regarding inference, it is the same as DETR trained normally, and only needs a single group of object queries.

The resulting architecture is equivalent to DETR with a group of parallel decoders, illustrated in Figure~\ref{fig:groupdetrandothers} (a). During training, the parallel decoders boost each other through sharing decoder parameters and using different object queries. On the other hand, using more groups of object queries resembles data augmentation, and behaves as query augmentation. It introduces more supervision and improve the decoder training. In addition, it is empirically observed that the encoder training is also improved, presumably with the help of the improved decoder.

Group DETR is versatile and is applicable to various DETR variants. Extensive experiments demonstrate that our approach is effective in achieving fast training convergence, shown in Figure~\ref{fig:curves}. Group DETR obtains consistent improvements on various DETR-based methods~\cite{meng2021conditional,liu2022dab,li2022dn,zhang2022dino}. For instance, Group DETR significantly improves Conditional DETR-C$5$ by ${5.0}$ mAP with $12$-epoch training on COCO~\cite{lin2014coco}. The non-trivial improvements hold when we adopt longer training schedules ({e.g.}, $36$ epochs and $50$ epochs). Furthermore, Group DETR outperforms baseline methods for multi-view $3$D object detection~\cite{liu2022petr,liu2022petrv2} and instance segmentation~\cite{cheng2021masked}.

\section{Background}

\noindent\textbf{DETR Architecture.} DETR~\cite{detr} is composed of an encoder, a transformer decoder, and object class and box position predictors. The encoder takes an image $\mathbf{I}$ as input, and outputs the image feature $\mathbf{X}$,
\begin{align}
\operatorname{Encoder}(\mathbf{I})   \rightarrow \mathbf{X}.
\end{align}
The decoder receives the image feature $\mathbf{X}$ and the \emph{object queries}, denoted by a matrix $\mathbf{Q}$$(=[\mathbf{q}_1~\mathbf{q}_2~\dots\mathbf{q}_N])$ as input, and outputs the embeddings $\tilde{\mathbf{Q}}$, followed by the predictors with the output denoted by $\mathbf{Y}$$(=[\mathbf{y}_1~\mathbf{y}_2~\dots\mathbf{y}_N])$,
\begin{align}
\operatorname{Decoder}(\mathbf{X}, \mathbf{Q}) \rightarrow \tilde{\mathbf{Q}},~~\operatorname{Predictor}(\tilde{\mathbf{Q}}) \rightarrow \mathbf{Y}. \label{eqn:singledecoder}
\end{align}

The decoder is a sequence of multiple layers. Each layer includes: (i) self-attention over object queries, which performs interactions among queries for collecting the information about {duplicate detection}; (ii) cross-attention between queries and image features, which collects the information from image features that is useful for object detection; (iii) feed-forward network that processes the queries separately to benefit object detection.

\vspace{1mm}
\noindent\textbf{DETR Training.} The predictions during DETR training are in the set form, and have no correspondence to the ground-truth objects. DETR uses one-to-one assignment, i.e., one ground-truth object is assigned to one predictions and vice versa, through building a bipartite matching between the predictions and the ground-truth objects:
\begin{align}(\mathbf{y}_{\sigma(1)}, \bar{\mathbf{y}}_1), 
(\mathbf{y}_{\sigma(2)}, \bar{\mathbf{y}}_2), 
\dots,
(\mathbf{y}_{\sigma(N)}, \bar{\mathbf{y}}_N).
\end{align}
Here, $\sigma(\cdot)$ is the optimal permutation of $N$ indices, and $[\bar{\mathbf{y}}_1~\bar{\mathbf{y}}_2, \dots~\bar{\mathbf{y}}_N] = \bar{\mathbf{Y}}$ correponds to ground truth. The loss is then formulated as below:
\begin{align}
    \mathcal{L} = \sum\nolimits_{n=1}^N\ell(\mathbf{y}_{\sigma(n)}, \bar{\mathbf{y}}_n),
\end{align}
where $\ell(\cdot)$ is a combination of the classification loss and the box regression loss between the ground-truth object $\bar{\mathbf{y}}$ and the prediction $\mathbf{y}$~\cite{detr,deformable-detr,meng2021conditional}.

Optimization with one-to-one assignment aims to score the predictions for promoting one prediction for one ground-truth object, and demoting duplicate predictions. Such scoring needs the comparison of one prediction with other predictions, and the information of other predictions is provided from decoder self-attention over queries. The two designs, \emph{one-to-one assignment and self-attention over object queries}, are critical for end-to-end detection without the need of the post-processing NMS.

\vspace{1mm}
\noindent\textbf{One-to-many assignment for non-end-to-end detection.} One-to-many assignment is successfully adopted for introducing more supervision to non-end-to-end detection training, such as Faster R-CNN~\cite{ren2015faster}, FCOS~\cite{tian2019fcos}, and so on~\cite{he2017mask,lin2017focal,redmon2018yolov3,zhang2020bridging,ge2021ota,chen2021you,ge2021yolox}. One ground-truth object is assigned to multiple anchors or multiple pixels. During inference, a post-processing NMS is conducted for duplicate detection removal.

\section{Group DETR}
\subsection{Algorithm}
\noindent\textbf{Naive one-to-many assignment.} We start from a naive way for one-to-many assignment depcited in Figure~\ref{fig:groupdetrandothers} (c). We replace one-to-one assignment with one-to-many assignment: assign one ground-truth object to multiple predictions. It does not work and the performance is much low. The reason is that the model is trained to output multiple predictions for one ground-truth object, and lacks the scoring mechanism to promote one single prediction and demote duplicate predictions for one ground-truth object.
\vspace{1mm}
\begin{algorithm}\small
\caption{Pseudocode of one Group Decoder Layer}
\label{alg:groupdecoder}
\definecolor{codeblue}{rgb}{0.25,0.5,0.5}
\definecolor{codekw}{rgb}{0.85, 0.18, 0.50}
\lstset{
  backgroundcolor=\color{white},
  basicstyle=\fontsize{7.5pt}{7.5pt}\ttfamily\selectfont,
  columns=fullflexible,
  breaklines=true,
  captionpos=b,
  commentstyle=\fontsize{7.5pt}{7.5pt}\color{codeblue},
  keywordstyle=\fontsize{7.5pt}{7.5pt}\color{codekw},
}
\begin{lstlisting}[language=python]
# SA: Self-Attention in the decoder layer
# CA: Cross-Attention in the decoder layer
# FFN: FFN in the decoder layer
# X: output image features of the encoder
# Q: object queries, with size (KxN, B, C)
# N, K, B, C: object query number, group number, batch size, feature dimension

# group decoder
if training:
    # split object queries to K groups
    Q_list = Q.split(N, dim=0) # a list of K tensors
    parallel_Q = cat(Q_list, dim=1) # (N, KxB, C)
    
    # parallel self-attention
    out = SA(parallel_Q) # (N, KxB, C)
    # concat all groups: (KxN, B, C)
    out = cat(out.split(B, dim=1), dim=0)
    
    # cross-attention and ffn
    out = FFN(CA(out, X))
else:   
    # in inference, only one group is kept
    Q = Q[:N]  # (N, B, C)
    
    # self-attention, cross-attention, and ffn
    out = SA(Q)
    out = FFN(CA(out, X))
\end{lstlisting}
\vspace{-3mm}
\end{algorithm}

\noindent\textbf{Group-wise one-to-many assignment.} We adopt the multi-group object query mechanism: form the initial $N$ queries as the primary group and introduce more $(K-1)$ groups of $N$ queries, totally $K$ groups, $\{\mathbf{Q}_1, \mathbf{Q}_2,\dots, \mathbf{Q}_K\}$. Accordingly, we have $K$ groups of predictions, $\{\mathbf{Y}_1, \mathbf{Y}_2,\dots, \mathbf{Y}_K\}$. We perform one-to-one assignment for each group, and find a bipartite matching $\sigma_k(\cdot)$, between each group of predictions and the ground-truth objects $(\mathbf{Y}_k, \bar{\mathbf{Y}})$. This results in that only one prediction for one ground-truth object is expected to score higher, and duplicate predictions is expected to score lower within one group other than within all the groups.

\vspace{1mm}
\noindent\textbf{Separate self-attention.} One-to-one assignment in one group means that the prediction assigned to one ground-truth object is superior to other predictions within the same group. This implies that we only need to collect the information of the predictions only from the same group, {rather than from all the groups}. Thus we perform self-attention (abbreviated as $\operatorname{SA}$) over queries for each group separately:
\begin{align}
\operatorname{SA(\mathbf{Q}_1)},
\operatorname{SA(\mathbf{Q}_2)},
\dots,
\operatorname{SA(\mathbf{Q}_K)}.
\end{align}

\vspace{1mm}
\begin{figure*}
\centering
\includegraphics[width=1.\textwidth]{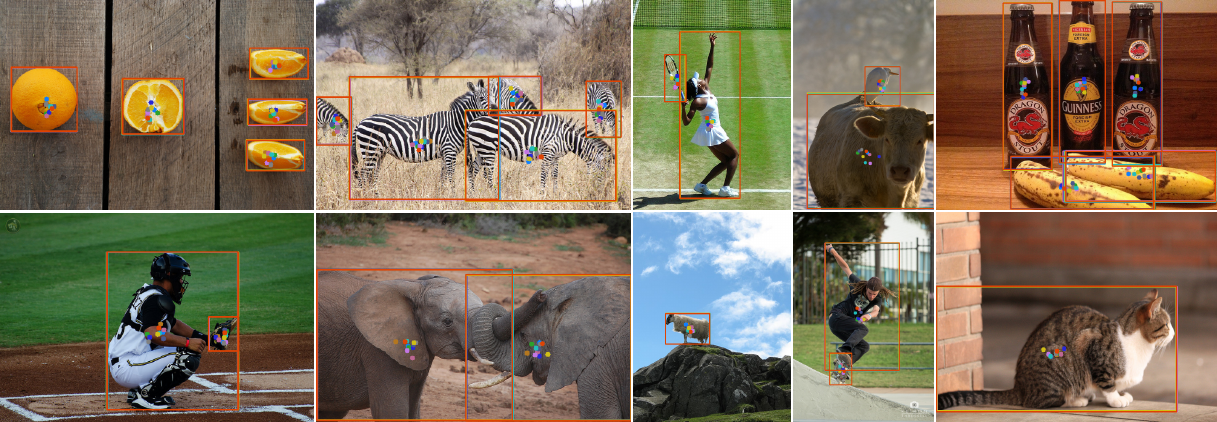}
 \caption{\textbf{Illustrating object queries.} The predicted boxes and reference points corresponding to object queries in different groups for the same ground-truth object are plotted in different colors with one color for one group. It can be seen that these queries are spatially close and can be viewed as an augmentation of other queries. The results are from Group DETR over Conditional DETR-R50~\cite{meng2021conditional}. The predicted boxes and reference points may overlap. Best view in color and zoom in.}
\label{fig:queryaugmentationillustration}
\end{figure*}
\begin{figure}
\centering
\begin{tikzpicture}[font=\footnotesize]
\pgfplotsset{set layers, compat=1.8, 
every axis/.append style={
font=\scriptsize,
}
}
    \begin{groupplot}[
        group style={
            group size=1 by 2,
            vertical sep=0pt
        },
        axis lines = left,
        every outer y axis line/.style={draw=none},
        every outer x axis line/.style={draw=gray!50},
        tick style={draw=none},
        yticklabels={},
        xticklabels={},
    	footnotesize,
        scale only axis,
    	x post scale=1.5,
    	xmin=0.2,
    	xmax=11.9
    ]
        \nextgroupplot[
          ybar,
          bar width=10pt,
          y post scale=0.4,
          ymin=35.0,
          ymax=37.65,
          enlargelimits=false,
    	  enlarge x limits=0.05,
    	  extra tick style={grid=major},
          grid style={line width=.1pt, draw=gray!20}
        ]
    
        \addplot[fill=mediumelectricblue, draw=mediumelectricblue, fill opacity=0.5, draw opacity=0.5, text opacity=1] 
	coordinates{
	(1, 37.6)
	(2, 37.5)
	(3, 37.4)
	(4, 37.5)
	(5, 37.6)
	(6, 37.5)
	(7, 37.6)
	(8, 37.4)
	(9, 37.5)
	(10, 37.6)
	(11, 37.5)
	};
        \draw[lightsalmon!60, dashed] (axis cs:0.7,37.4) -- (axis cs:11.5,37.4);
        \draw[lightsalmon] (axis cs:-0.4,37.5) node[right]{37.5} (axis cs:0.7,37.5) -- (axis cs:11.5,37.5);
        \draw[lightsalmon] (axis cs:11.4,37.5) node[right]{$\pm$0.1};
        \draw[lightsalmon!60, dashed] (axis cs:0.7,37.6) -- (axis cs:11.5,37.6);
    \end{groupplot}
\end{tikzpicture}
\caption{\textbf{The performance across groups of queries are similar.} Only a $\pm$$0.1$ mAP is observed over the median ($37.5$ mAP). The mAP scores over the COCO {\em val2017} are reported by a $12$-epoch trained Conditional DETR-R50 with Group DETR.}
\label{fig:similarperformanceacrossgroups}
\vspace{-2mm}
\end{figure}
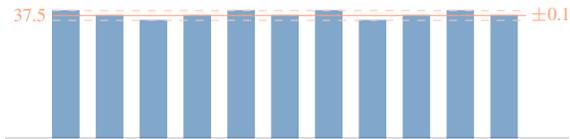

\noindent\textbf{Training architecture.} The resulting architecture for training is very simple: the encoder keeps the same, and the decoder contains $K$ separate parallel decoders as shown in Figure~\ref{fig:groupdetrandothers} (a):
\begin{align}
\operatorname{Decoder}(\mathbf{X}, \mathbf{Q}_1)   \rightarrow \mathbf{Q}_1,~&\operatorname{Predictor}(\mathbf{Q}_1)
\rightarrow \mathbf{Y}_1, \nonumber \\
\operatorname{Decoder}(\mathbf{X}, \mathbf{Q}_2)   \rightarrow \mathbf{Q}_2,~~&\operatorname{Predictor}(\mathbf{Q}_2)
\rightarrow \mathbf{Y}_2, \nonumber\\
\dots ~& \dots \nonumber\\
\operatorname{Decoder}(\mathbf{X}, \mathbf{Q}_K)   \rightarrow \mathbf{Q}_K,~&\operatorname{Predictor}(\mathbf{Q}_K)
\rightarrow \mathbf{Y}_K. \label{eqn:groupdecoder}
\end{align}
Here, the parameters of the decoder and the predictor for the $K$ groups are shared. Decoder separation and parallelism are feasible in that there is no interaction among queries for the other two operations, cross-attention and FFN. Our approach is called Group Decoder. In model inference, the process is the same as DETR trained normally and only needs one group of queries without any architecture modification. The pseudo-code is shown in Algorithm~\ref{alg:groupdecoder}.

\vspace{1mm}
\noindent\textbf{Loss function.} The loss is an aggregation of $K$ losses, each for one decoder. It is written as follows,
\begin{align}
    \mathcal{L} = \frac{1}{K}\sum_{k=1}^K \mathcal{L}_k
    = \frac{1}{K} \sum_{k=1}^K\sum_{n=1}^N\ell(\mathbf{y}_{\sigma_k(n)}, \bar{\mathbf{y}}_{kn}),
    \label{eqn:groupdetrloss}
\end{align}
where $\sigma_k(\cdot)$ is the optimal permutation of $N$ indices for the $k$th decoder.

\subsection{Analysis} \label{subsec:analysis}

\noindent\textbf{Explanation with parameter-shared models.} We discuss Group DETR from the perspective of training multiple models with parameter sharing. Training with Group DETR can be regarded as simultaneously training $K$ DETR models, which share the parameters of the encoder, the decoder, and the predictor, and only differ in the initialization of object queries. This leads to the shared parameters receive more back-propagated gradients. Thus, these parameters are better trained and accordingly the training process converges faster.

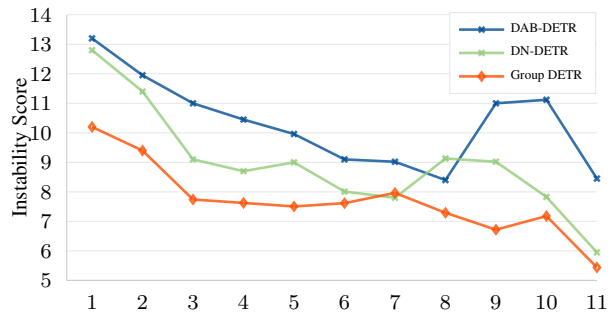
\begin{figure}[t]
\centering
\begin{tikzpicture}[font=\footnotesize]
\begin{axis}[
legend columns=1, 
legend style={at={(0.86,1.01)},anchor=north, font=\tiny, draw=gray!20}, legend cell align={left},
y label style={at={(-0.05,0.5)}},
ylabel={Instability Score}, ymajorgrids=true, xtick={1,2,3,4,5,6,7,8,9,10,11}, xticklabels={$1$, $2$, $3$, $4$, $5$, $6$, $7$, $8$, $9$, $10$, $11$},
ytick={5, 6,7,8,9,10,11,12,13, 14},
tick style={draw=none},
y label style={font=\footnotesize},
height=6cm,
width=9cm,
y post scale=0.8,
x post scale=0.95,
axis lines = left,
every outer y axis line/.style={draw=gray!40},
every outer x axis line/.style={draw=gray!40},
grid style={line width=.1pt, draw=gray!20},
xmin=0.5, ymax=14, ymin=5]

\addplot[line width=1.0pt, mark size=1.5pt, mark=x, draw=mediumelectricblue, draw opacity=0.8]
table
{
X Y
1 13.2
2 11.95
3 11.0
4 10.45
5 9.96
6 9.1
7 9.02
8 8.4
9 11.0
10 11.12
11 8.45
};

\addplot[line width=1.0pt, mark size=1.5pt, mark=x, draw=mossgreen2, draw opacity=0.8]
table
{
X Y
1 12.8
2 11.4
3 9.1
4 8.7
5 9.0
6 8.01
7 7.8
8 9.13
9 9.02
10 7.83
11 5.95
};

\addplot[line width=1.0pt, mark size=1.5pt, mark=diamond, draw=internationalorange, draw opacity=0.8]
table
{
X Y
1 10.200400352478027
2 9.398599624633789
3 7.740799903869629
4 7.6255998611450195
5 7.5055999755859375
6 7.617400169372559
7 7.964799880981445
8 7.290999889373779
9 6.719600200653076
10 7.1793999671936035
11 5.440199851989746
};
\addlegendentry{DAB-DETR}   
\addlegendentry{DN-DETR}
\addlegendentry{Group DETR}  
\end{axis}
\end{tikzpicture}
\caption{\textbf{More stable assignment.} The $x$-axis corresponds to \#epoch, and the $y$-axis corresponds to instability score (the score is introduced by DN-DETR~\cite{li2022dn}, the lower the instability score, the more stable the label assignment) over COCO {\em val2017}. One can see that the assignment in Group DETR is more stable than DN-DETR and its baseline DAB-DETR.}
\label{fig:onetooneassignmentstablization}
\end{figure}

As a side benefit, we observe that Group DETR makes the assignment more stable, as shown in Figure~\ref{fig:onetooneassignmentstablization}. We speculate that the stability is because the improved network leads to more reliable predictions, and thus the assignment quality is better.

\vspace{1mm}
\noindent\textbf{Explanation with object query augmentation.} The multi-group object query mechanism introduces additional $(K-1)$ group of queries, which can be regarded as an augmentation of the primary group of queries. This is empirically illustrated in Figure~\ref{fig:queryaugmentationillustration}. The reference points predicting the same objects are spatially close, and thus the corresponding object queries are similar. This may suggest that the multi-group object query mechanism resembles data augmentation, and at each iteration, more automatically-learned augmented queries are included, which equivalently introduces more supervision for decoder training. The results in Figure~\ref{fig:similarperformanceacrossgroups} empirically suggest that different groups of augmented queries lead to similar results.

The point about more supervision is also observed from the comparison between Equation~\ref{eqn:groupdecoder} (for training with Group DETR) and Equation~\ref{eqn:singledecoder} (for normal DETR training). Group DETR training includes $K$ pairs of image feature and object query group $\{(\mathbf{X}, \mathbf{Q}_1), (\mathbf{X}, \mathbf{Q}_2), \dots, (\mathbf{X}, \mathbf{Q}_K)\}$, and thus the loss contains more components as shown in Equation~\ref{eqn:groupdetrloss}.

\begin{table}[h]
  \centering
    \setlength{\tabcolsep}{7pt}
    \renewcommand{\arraystretch}{1.25}
    \footnotesize
      \caption{\textbf{Illustrating that training with Group DETR improves both encoder and decoder.} The encoder, including CNN and transformer encoders, is initialized from a trained Conditional DETR-R50~\cite{meng2021conditional} with $50$ epochs and the decoder is random initialized. (a) (Fixed, Single) = the encoder is not retrained, and the decoder is trained normally without using Group DETR. (b) (Fixed, Group) = the encoder is not retrained, and the decoder is with Group DETR. (c) (Group, Group) = the encoder and the decoder are trained with Group DETR. All the results are got through training with $50$ epochs. (c) $>$ (b) implies that Group DETR also improves the encoder training.}
  \label{tab:benefittoencoderdecoder}
  \vspace{2mm}
  \begin{tabular}{ccc|cccc}
    \shline
    & Encoder & Decoder & mAP & AP$_{s}$ & AP$_{m}$ & AP$_{l}$ \\
    \hline
    (a) & Fixed & Single & $40.6^*$ & $20.2$ & $44.0$ & $59.3$ \\
    (b) & Fixed & Group & $41.5$ & $21.2$ & $45.0$ & $60.2$ \\
    (c) & Group & Group & $42.9^*$ & $22.2$ & $46.6$ & $61.6$ \\
    \shline
  \end{tabular}
  \\
  $^*$: Training Conditional DETR with a trained encoder gives slightly lower performances than the one trained regularly, even though we train all components. New hyper-parameters may need to get better results.
\vspace{-1mm}
\end{table}

\vspace{1mm}
\noindent\textbf{Encoder training improvement.} The additional supervision introduces more box regression and classification supervision from more queries assigned to each ground-truth object. The gradients with more supervision are also back-propagated from the decoder to the encoder. It is presumable that the encoder also gets benefit, verified by the empirical results in Table~\ref{tab:benefittoencoderdecoder}.

\begin{figure}
    \centering    
    \includegraphics{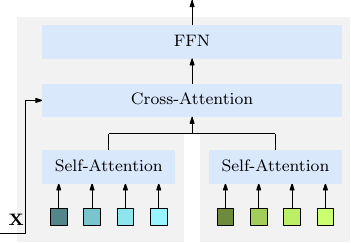}
    \caption{The parallel decoders in Group DETR are efficiently implemented as parallel self-attention, cross attention and FFN.}
    \label{fig:GroupDETRwithParallelSelfAttention}
    \vspace{-3mm}
\end{figure}
\begin{figure}
\centering
\begin{tikzpicture}[font=\footnotesize]
\pgfplotsset{compat=1.11,
    /pgfplots/xbar legend/.style={
    /pgfplots/legend image code/.code={%
       \draw[##1,/tikz/.cd,yshift=-0.25em]
        (0cm,0cm) rectangle (7pt,0.8em);},
   },
}
\begin{axis}[xbar,
             bar width=10pt,
             xmin=0,
             xmax=7.0,
             ymin=-1.0,
             ymax=1.6,
             ytick=data,
             ytick pos=left,
             nodes near coords,
             yticklabels={DAB-DETR, C-DETR},
             reverse legend,
             legend  style={at={(0.26, 0.18)}, anchor=north, legend columns =-1, draw=none, fill=none, font=\tiny},
             axis line style={draw=none},
             tick style={draw=none},
             xticklabels={},
             y post scale=0.62,
             x post scale=0.95
             ]
\addplot[fill=mossgreen, draw=mossgreen, fill opacity=0.8, draw opacity=0.8] table[y expr=\coordindex, x index=0,row sep=\\] {4.8\\4.5\\};
\addplot[fill=mediumelectricblue, draw=mediumelectricblue, fill opacity=0.5, draw opacity=0.5] table[y expr=\coordindex, x index=0,row sep=\\] {3.1\\3.3\\};
\addplot[fill=mediumelectricblue, draw=mediumelectricblue, fill opacity=0.1, draw opacity=0.1] table[y expr=\coordindex, x index=0,row sep=\\] {6.5\\5.8\\};
\legend{Group DETR, Baseline, Baseline}
\end{axis}
\end{tikzpicture}
\caption{\textbf{Baseline models {vs} their Group DETR counterparts w.r.t training memory}. The {\color{mediumelectricblue!20} gray} baseline represents using the naive implementation of attention modules. With a memory-efficient implementation~\cite{dao2022flashattention}, Group DETR does not bring much memory burden during training, only requires $1.2$ G and $1.7$ G more GPU memory with Conditional DETR~\cite{meng2021conditional} (`C-DETR' for short) and DAB-DETR~\cite{liu2022dab}.}
\label{fig:memorycostcomparison}
\end{figure}
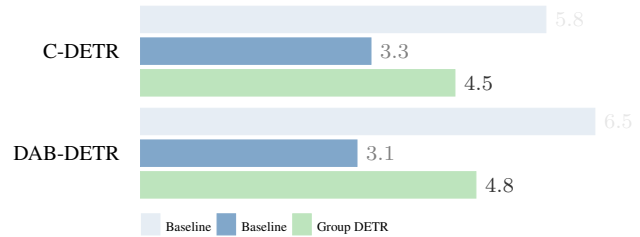
\begin{table}[t]
  \centering
    \setlength{\tabcolsep}{6pt}
    \renewcommand{\arraystretch}{1.25}
    \footnotesize
      \caption{\textbf{Group DETR outperforms baseline models with a similar training time.} Conditional DETR~\cite{meng2021conditional} and DAB-DETR~\cite{liu2022dab} serve as baseline models to compare the performances on COCO {\em val2017}~\cite{lin2014coco}. `C-DETR' and `w/ Group' are the abbreviations of `Conditional DETR' and `with Group DETR'. The entries noted by {\color{gray!50} grey} are the results of baseline models with the same training epochs ($12$ or $50$ epochs) as Group DETR. To match the training times of Group DETR, we adopt longer training shedules for baselines ($15$ or $60$ epochs). The training times are measured on $8$ A$100$ GPUs in hours.}
  \label{tab:comparabletrainingtime}
  \vspace{2mm}
  \begin{tabular}{l|c|ccccc}
    \shline
    Model & w/ Group & Hours & mAP & AP$_{s}$ & AP$_{m}$ & AP$_{l}$ \\
    \hline
    \multirow{3}{*}{C-DETR} &  & \color{gray!50}{$4.6$} & \color{gray!50}{$32.6$} & \color{gray!50}{$14.7$} & \color{gray!50}{$35.0$} & \color{gray!50}{$48.3$}  \\
    &  & $5.8$ & $34.4$ & $15.1$ & $37.3$ & $51.3$ \\
   &\checkmark & $5.6$ & $37.6$ & $18.2$ & $40.7$ & $55.9$ \\
   \hline
   \multirow{3}{*}{C-DETR} &  & \color{gray!50}{$19.2$} & \color{gray!50}{$40.9$} & \color{gray!50}{$20.5$} & \color{gray!50}{$44.2$} & \color{gray!50}{$59.6$} \\
    &  & $23.0$ & $41.6$ & $21.4$ & $45.1$ & $60.0$ \\
    & \checkmark & $23.3$ & $43.4$ & $23.0$ & $47.3$ & $62.3$ \\
    \hline
    \multirow{3}{*}{DAB-DETR} & & \color{gray!50}{$5.6$} & \color{gray!50}{$35.2$} & \color{gray!50}{$16.7$} & \color{gray!50}{$38.6$} & \color{gray!50}{$51.6$} \\
    & & $7.0$ & $36.3$ & $17.1$ & $39.4$ & $52.5$ \\
   & \checkmark & $6.6$ & $39.1$ & $19.7$ & $42.5$ & $56.8$ \\
   \hline
   \multirow{3}{*}{DAB-DETR} & & \color{gray!50}{$23.3$} & \color{gray!50}{$42.2$} & \color{gray!50}{$21.5$} & \color{gray!50}{$45.7$} & \color{gray!50}{$60.3$} \\
    & & $28.0$ & $42.9$ & $22.8$ & $46.4$ & $61.9$ \\
   & \checkmark & $27.5$ & $44.5$ & $24.2$ & $48.5$ & $63.2$ \\
   \shline
  \end{tabular}
  \vspace{-2mm}
\end{table}
\vspace{1mm}

\noindent\textbf{Computation and memory complexity.} Group DETR uses more decoders during training. It is expected that Group DETR will bring additional training computation costs (FLOPs) as well as training memory costs. But the parallel decoders can be implemented as a single decoder by replacing normal self-attention with parallel self-attention (depicted in Figure~\ref{fig:GroupDETRwithParallelSelfAttention}) and we can use an efficient attention implementation, FlashAttention~\cite{dao2022flashattention,xFormers2022}. As a result, Group DETR only takes a small increase in training GPU memory and training time. For example, with Conditional DETR~\cite{meng2021conditional} and DAB-DETR~\cite{liu2022dab}, the memory increases are just $1.2$ G and $1.7$ G (Figure~\ref{fig:memorycostcomparison}). The training time is increased by $5$ minutes per epoch (from $23$ minutes to $28$ minutes and from $28$ minutes to $33$ minutes, respectively).

We provide the results by increasing the training time for normal DETR training to see if Group DETR benefits simply from more training time. The results given in Table~\ref{tab:comparabletrainingtime} show that normal training with more training time brings a little benefit and the performance is still much lower than Group DETR, implying that the performance gain from our approach is not from training time increase. 

\vspace{1mm}
\noindent\textbf{Connection to DN-DETR.} DN-DETR~\cite{li2022dn} aims to stabilize one-to-one assignment during DETR training. DN-DETR forms the additional queries by adding the noises to ground-truth objects, which can be regarded as a variant of our multi-group mechanism with clear differences. In DN-DETR~\cite{li2022dn}, on the one hand, the number of queries within each additional group is the same as the number of ground-truth objects. Each one correspond to one ground-truth object, and there is no query corresponding to no-object. In contrast, our approach automatically learns a number of $N$ (e.g., $300$) object queries that correspond to both ground-truth objects and no-object.

On the other hand, DN-DETR performs self-attention over noised queries, mainly for collecting the information from predictions for other objects other than from duplicate predictions. Self-attention in Group DETR instead collects both duplicate predictions and predictions for other objects.

The above two comparisons imply that DN-DETR brings the major help for the box and classification prediction, through the introduction of more positive queries corresponding to ground-truth objects (like FCOS), and no direct help for duplicate prediction removal. Our approach introduces both positive queries and negative queries (no-object), also brings the help for duplicate prediction removal.

\begin{figure}
\centering
\begin{tikzpicture}[font=\footnotesize]
\pgfplotsset{compat=1.11,
    /pgfplots/xbar legend/.style={
    /pgfplots/legend image code/.code={%
       \draw[##1,/tikz/.cd,yshift=-0.25em]
        (0cm,0cm) rectangle (7pt,0.8em);},
   },
}
\begin{axis}[xbar,
             bar width=10pt,
             xmin=32,
             xmax=42,
             ymin=-0.5,
             ymax=0.5,
             ytick=data,
             ytick pos=left,
             nodes near coords,
             yticklabels={DAB-DETR},
             yticklabel style = {rotate=90},
             enlarge x limits={value=0.2,upper},
             grid style={line width=.1pt, draw=gray!20},
             axis line style={draw=none},
             tick style={draw=none},
             xticklabels={},
             y post scale=0.4,
             x post scale=1.28,
             reverse legend,
             legend cell align=left,
                legend style={
                        draw=none,
                        fill=none,
                        at={(0.341,0.1)},
                        anchor=north,
                        legend columns=-1,
                        font=\tiny,
                }
             ]
\addplot[fill=lightsalmon, draw=lightsalmon, fill opacity=0.6, draw opacity=0.6] table[y expr=\coordindex, x index=0,row sep=\\] {40.6\\};
\addplot[fill=mossgreen, draw=mossgreen, fill opacity=0.8, draw opacity=0.8] table[y expr=\coordindex, x index=0,row sep=\\] {39.1\\};
\addplot[fill=babyblue, draw=babyblue, fill opacity=0.8, draw opacity=0.8] table[y expr=\coordindex, x index=0,row sep=\\] {38.8\\};
\addplot[fill=mediumelectricblue, draw=mediumelectricblue, fill opacity=0.5, draw opacity=0.5] table[y expr=\coordindex, x index=0,row sep=\\] {35.2\\};
\legend{DN-DETR + Group DETR, Group DETR, DN-DETR$^*$, Baseline}
\end{axis}
\end{tikzpicture}
\vspace{1mm}
\caption{\textbf{Comparisons with DN-DETR.} Group DETR outperforms DN-DETR on DAB-DETR~\cite{liu2022dab} ($y$-axis). Combining those two methods give better results, indicating they are complementary to each other. The $x$-axis is the mAP scores with a $12$-epoch schedule on COCO {\em val2017}. $^*$ represents that we report the best results of DN-DETR among different numbers of denoising queries (detailed results are provided in Appendix).}
\label{fig:comparewithdndetr}
\vspace{-2mm}
\end{figure}
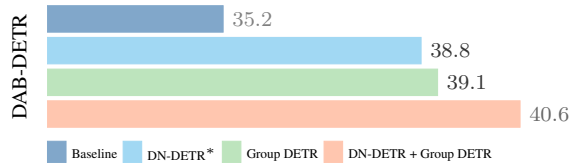

Figure~\ref{fig:comparewithdndetr} shows that the performance of Group DETR is better than DN-DETR. We further investigate if Group DETR still benefits from introducing more positive queries with noised queries. As shown in Figure~\ref{fig:comparewithdndetr}, the performance gain over Group DETR is non-trivial, a $1.5$ mAP. This implies that Group DETR and DN-DETR are complementary and their major roles are different, though they have some similarities.

\section{Experiments}
We demonstrate the effectiveness of Group DETR in various DETR variants, and its extension to $3$D detection and instance segmentation~\cite{meng2021conditional,liu2022dab,li2022dn,deformable-detr,zhang2022dino,liu2022petr,liu2022petrv2,cheng2021masked}. The training setting is almost the same as baseline models, for illustrating the effectiveness of our Group DETR. We adopt the same training settings and hyper-parameters as the baseline models, such as learning rate, optimizer, pre-trained model, initialization methods, and data augmentations\footnote{We may adjust the batch size due to the limitation of the GPU memory size for both the baseline model and our approach so that the batch size is the same.}.

\subsection{Object Detection}

\noindent\textbf{Setting.} We study various representative DETR-based detectors, such as basic baselines (Conditional DETR~\cite{meng2021conditional}, DAB-DETR~\cite{liu2022dab}, DN-DETR~\cite{li2022dn}) with dense attentions, and strong baselines (DAB-Deformable-DETR~\cite{liu2022dab,deformable-detr} and DINO~\cite{zhang2022dino,deformable-detr}) with deformable attentions. We report the results on two training schedules, training for $12$ epochs and training for more epochs ($36$ or $50$). Unless specified, the models are trained with ResNet-50~\cite{he2016deep} as the backbone on the COCO {\em train2017} and evaluated on the COCO {\em val2017}. More implementation details are provided in Appendix.

\begin{table}
  \centering
    \setlength{\tabcolsep}{4pt}
    \renewcommand{\arraystretch}{1.5}
    \footnotesize
  \caption{\textbf{Effectiveness of Group DETR with $12$ epochs.} Group DETR gives consistent gains over various DETR-based baselines on COCO {\em val2017}~\cite{lin2014coco}, highlighted with brackets. All experiments adopt ResNet-50~\cite{he2016deep} and do not use multiple patterns~\cite{wang2021anchor}. For DN-DETR, an improved version of DN, dynamic DN groups~\cite{zhang2022dino} with 100 DN queries, is used, making the results slightly different from the ones (with $3$ patterns) reported in the original paper~\cite{li2022dn} (more results about the number of DN queries can be found in Appendix. `C-DETR', `DAB-D-DETR', and `w/ Group' are `Conditional DETR'~\cite{meng2021conditional}, `DAB-Deformable DETR'~\cite{liu2022dab,deformable-detr}, and `with Group DETR', respectively, for neat representation.}
  \vspace{2mm}
  \begin{tabular}{l|c|cccc}
    \shline
    Model     & w/ Group & mAP & AP$_s$ & AP$_m$ & AP$_l$ \\
    \hline
    \multirow{2}{*}{C-DETR} &  & $32.6$ & $14.7$ & $35.0$ & $48.3$     \\
     & \checkmark & \reshl{$37.6$}{+}{$\mathbf{5.0}$} & $18.2$ & $40.7$ & $55.9$ \\
     \hline
    \multirow{2}{*}{C-DETR-DC$5$} &  & $36.4$ & $18.0$ & $39.6$ & $52.5$     \\
     & \checkmark & \reshl{$41.2$}{+}{$\mathbf{4.8}$} & $21.4$ & $45.0$ & $58.7$ \\
    \hline
    \multirow{2}{*}{DAB-DETR} &  & $35.2$ & $16.7$ & $38.6$ & $51.6$     \\
     & \checkmark & \reshl{$39.1$}{+}{$\mathbf{3.9}$} & $19.7$ & $42.5$ & $56.8$ \\
     \hline
    \multirow{2}{*}{DAB-DETR-DC$5$} &  & $37.5$ & $19.4$ & $40.6$ & $53.2$     \\
     & \checkmark & \reshl{$41.9$}{+}{$\mathbf{4.4}$} & $23.3$ & $45.6$ & $58.4$ \\
    \hline
    \multirow{2}{*}{DN-DETR} &  & $38.6$ & $17.9$ & $41.6$ & $57.7$     \\
     & \checkmark & \reshl{$40.6$}{+}{$\mathbf{2.0}$} & $19.8$ & $43.9$ & $59.4$ \\
     \hline
    \multirow{2}{*}{DN-DETR-DC$5$} &  & $41.9$ & $22.2$ & $45.1$ & $59.8$     \\
     & \checkmark & \reshl{$44.5$}{+}{$\mathbf{2.6}$} & $25.9$ & $48.2$ & $62.2$ \\
    \hline
    \multirow{2}{*}{DAB-D-DETR} &  & $44.2$ & $27.5$ & $47.1$ & $58.6$     \\
     & \checkmark & \reshl{$45.7$}{+}{$\mathbf{1.5}$} & $28.1$ & $49.0$ & $60.6$ \\
    \hline
    \multirow{2}{*}{DINO-$4$scale} &  & $49.4$ & $32.3$ & $52.5$ & $63.2$     \\
     & \checkmark & \reshl{$\mathbf{50.1}$}{+}{$\mathbf{0.7}$} & $32.4$ & $53.2$ & $64.7$ \\
    \shline
  \end{tabular}
  \label{tab:main1}
  \vspace{-2mm}
\end{table}
\begin{table}[h]
  \centering
    \setlength{\tabcolsep}{3pt}
    \renewcommand{\arraystretch}{1.25}
    \footnotesize
  \caption{\textbf{Effectiveness of Group DETR with more epochs.} Group DETR still outperforms baselines by non-trivial margins with more training epochs ($36$ or $50$ epochs). Settings and notations are consistent with Table~\ref{tab:main1}, except for the training epochs ($36$ epochs for DINO-$4$scale by following the original paper~\cite{zhang2022dino} and $50$ epochs for other models). `DINO-$4$scale-Swin-L' means it adopts Swin-Large~\cite{liu2021swin} as the backbone.}
  \vspace{2mm}
  \begin{tabular}{l|c|cccc}
    \shline
    Model     & w/ Group & mAP & AP$_s$ & AP$_m$ & AP$_l$ \\
    \hline
    \multirow{2}{*}{C-DETR} &  & $40.9$ & $20.5$ & $44.2$ & $59.6$     \\
     & \checkmark & \reshl{$43.4$}{+}{$\mathbf{2.5}$} & $23.0$ & $47.3$ & $62.3$ \\
     \hline
    \multirow{2}{*}{C-DETR-DC$5$} &  & $43.7$ & $23.9$ & $47.6$ & $60.1$     \\
     & \checkmark & \reshl{$45.8$}{+}{$\mathbf{2.1}$} & $26.8$ & $49.7$ & $63.1$ \\
    \hline
    \multirow{2}{*}{DAB-DETR} &  & $42.2$ & $21.5$ & $45.7$ & $60.3$     \\
     & \checkmark & \reshl{$44.5$}{+}{$\mathbf{2.3}$} & $24.2$ & $48.5$ & $63.2$ \\
    \hline
    \multirow{2}{*}{DAB-DETR-DC$5$} &  & $44.5$ & $25.3$ & $48.2$ & $62.3$     \\
     & \checkmark & \reshl{$46.7$}{+}{$\mathbf{2.2}$} & $27.6$ & $50.9$ & $64.0$ \\
    \hline
    \multirow{2}{*}{DN-DETR} &  & $44.0$ & $23.9$ & $47.7$ & $62.9$     \\
     & \checkmark & \reshl{$45.4$}{+}{$\mathbf{1.4}$} & $25.1$ & $49.3$ & $63.8$ \\
    \hline
    \multirow{2}{*}{DN-DETR-DC$5$} &  & $47.5$ & $27.9$ & $50.7$ & $65.9$     \\
     & \checkmark & \reshl{$48.0$}{+}{$\mathbf{0.5}$} & $29.3$ & $52.1$ & $65.4$ \\
    \hline
    \multirow{2}{*}{DAB-D-DETR} &  & $48.1$ & $31.4$ & $51.4$ & $63.4$     \\
     & \checkmark & \reshl{$49.7$}{+}{$\mathbf{1.6}$} & $31.4$ & $52.5$ & $65.6$ \\
    \hline
    \multirow{2}{*}{DINO-$4$scale} &  & $50.9$ & $34.6$ & $54.1$ & $64.6$     \\
     & \checkmark & \reshl{$51.3$}{+}{$\mathbf{0.4}$} & $34.7$ & $54.5$ & $65.3$ \\
    \hline
    \multirow{2}{*}{DINO-$4$scale-Swin-L} &  & $58.0$ & $41.3$ & $61.9$ & $74.0$ \\
     & \checkmark & \reshl{$\mathbf{58.4}$}{+}{$\mathbf{0.4}$} & $41.0$ & $62.5$ & $73.9$ \\
    \shline
  \end{tabular}
  \label{tab:main2}
  \vspace{-3mm}
\end{table}

\vspace{1mm}
\noindent\textbf{Results.} We first report the results of training with $12$ epochs in Table~\ref{tab:main1}. Group DETR brings consistent improvements over the baselines with dense attentions that already are superior to the original DETR~\cite{detr}. It boosts Conditional DETR (-DC$5$)~\cite{meng2021conditional} by $\mathbf{5.0}$ ($\mathbf{4.8}$) mAP, improves DAB-DETR (-DC$5$)~\cite{liu2022dab} by $\mathbf{3.9}$ ($\mathbf{4.4}$) mAP, and brings a $\mathbf{2.0}$ ($\mathbf{2.6}$) mAP gain to DN-DETR (-DC$5$)~\cite{li2022dn}. 

Group DETR also works well on those strong baselines with deformable attentions that are equipped with two or more accelerating techniques. It gives a $\mathbf{1.5}$ mAP improvements over DAB-Deformable-DETR~\cite{liu2022dab,deformable-detr}. When applying to DINO~\cite{zhang2022dino,li2022dn,deformable-detr}, Group DETR also exceeds it by $0.7$ mAP. The gain is non-trivial over such a stronger baseline, considering that DINO is a well-tuned model\footnote{In fact, our approach is compatible with query denoising and two-stage. In Table $3$, for example, DN-DETR~\cite{li2022dn} utilizes query denoising, and our method improves it by $2.0$. Similarly, DAB-D-DETR~\cite{liu2022dab} adopts a two-stage structure, and our method achieves a $1.5$ improvement.} based on DAB-Deformable-DETR that combines improved hyper-parameters, improved two-stage design, improved query denoising task, and other tricks.

Furthermore, we report the results with $50$ training epochs that is commonly adopted in many acceleration methods~\cite{deformable-detr,meng2021conditional,chen2022conditional,liu2022dab}. Table~\ref{tab:main2} presents that Group DETR outperforms baseline models by large margins. For the stronger backbone, Swin-Large~\cite{liu2021swin}, our approach achieves $58.4$ mAP (still a $0.4$ mAP higher than its baseline DINO~\cite{zhang2022dino} ($58.0$ mAP with Swin-Large)). This verifies the generalization ability of our Group DETR.

Last, we compare the training convergence curves of the baseline models and their Group DETR counterparts. The results, as shown in Figure~\ref{fig:curves}, provide more evidence that Group DETR speeds DETR training convergence on various DETR variants.

\vspace{1mm}
\noindent\textbf{System-level Results on COCO {\em test-dev} with ViT-Huge.} We also have the system-level performance on COCO {\em test-dev}~\cite{lin2014coco} with ViT-Huge~\cite{dosovitskiy2020vit}. We apply Group DETR to DINO~\cite{zhang2022dino} and follow its training pipeline and settings: pretrain the encoder with a self-supervised method, then pretrain the whole model on Object365~\cite{shao2019objects365}, and last fine-tune the whole model on COCO~\cite{lin2014coco}. Our model is the first to achieve $\mathbf{64.5}$ mAP on COCO {\em test-dev}, which is still superior to other methods with larger encoder and more pre-training data~\cite{liu2022swin,wang2022image,wei2022contrastive,yang2022focal}. The details and comparisons with other methods are provided in Appendix.

\subsection{More Applications}
Group DETR is applicable to DETR-style techniques to other vision problems. We report the results for two additional problems: multi-view $3$D object detection~\cite{huang2021bevdet,li2022bevformer,liu2022petr,liu2022petrv2} and instance segmentation~\cite{cheng2021masked,li2022mask}, to further demonstrate the effectiveness.

\vspace{1mm}
\noindent\textbf{Multi-view 3D object detection.} We report the results over PETR~\cite{liu2022petr} and PETR v$2$~\cite{liu2022petrv2} on the nuScenes {\em val} dataset~\cite{caesar2020nuscenes}. Table~\ref{tab:bev_3d} shows that Group DETR brings significant gains to PETR and PETR v$2$ with $24$ training epochs in terms of both the nuScenes Detection Score (NDS) and mAP scores. 

\begin{table}[h]
  \centering
    \setlength{\tabcolsep}{3pt}
    \renewcommand{\arraystretch}{1.25}
    \footnotesize
  \caption{\textbf{Results on multi-view 3D object detection.} All experiments are evaluated on the nuScenes {\em val} set~\cite{caesar2020nuscenes}. We train these experiments for $24$ epochs with VoVNetV$2$~\cite{lee2020centermask} as the backbone and with the image size of $800\times 320$. We follow all the settings and hyper-parameters of PETR~\cite{liu2022petr} and PETR v$2$~\cite{liu2022petrv2}. }
  \vspace{2mm}
  \begin{tabular}{l|c|ccc}
    \shline
    Model     & w/ Group & NDS & mAP & \\
    \hline
    \multirow{2}{*}{PETR} & & $42.0$ & $37.4$ & \\
     & \checkmark & \reshl{$45.0$}{+}{$\mathbf{3.0}$} &   \reshl{$38.8$}{+}{$\mathbf{1.4}$}&\\
    \hline
    \multirow{2}{*}{PETR v$2$} & & $50.3$ & $40.7$ & \\
     & \checkmark & \reshl{$\mathbf{51.3}$}{+}{$\mathbf{1.0}$} & \reshl{$\mathbf{41.9}$}{+}{$\mathbf{1.2}$} & \\
    \shline
  \end{tabular}
  \vspace{-2mm}
  \label{tab:bev_3d}
\end{table}

\paragraph{Instance segmentation.} 
We demonstrate the effectiveness of the representative method, Mask2Former~\cite{cheng2021masked}. The results are given in Table~\ref{tab:mask2former}. Group DETR achieves a $1.2$ ($0.3$) mAP$^{m}$ gain with $12$ ($50$) epochs.

\begin{table}[h]
  \centering
    \setlength{\tabcolsep}{5pt}
    \renewcommand{\arraystretch}{1.25}
    \footnotesize
  \caption{\textbf{Results on instance segmentation.} The mask mAP (mAP$^{m}$) is used for instance segmentation on COCO {\em val2017}. We adopt Mask2Former~\cite{cheng2021masked} as the baseline. The experiments are conducted with ResNet-50~\cite{he2016deep} as the backbone, following all the settings of Mask2Former.}
  \vspace{3mm}
  \begin{tabular}{c|c|ccccc}
    \shline
    Epochs & w/ Group & mAP$^{m}$ & AP$_{s}^{m}$ & AP$_{m}^{m}$ & AP$_{l}^{m}$ \\
    \hline
     $12$ & & $38.5$ & $17.6$ & $41.4$ & $60.4$ \\
     $12$ & \checkmark & \reshl{$39.7$}{+}{$\mathbf{1.2}$} & $18.7$ & $42.8$ & $60.8$ \\
     \hline
     $50$ & & $43.7$ & $23.4$ & $47.2$ & $64.8$ \\
     $50$ &  \checkmark & \reshl{$\mathbf{44.0}$}{+}{$\mathbf{0.3}$} & $23.8$ & $47.1$ & $65.1$ \\
    \shline
  \end{tabular}
  \vspace{-2mm}
  \label{tab:mask2former}
\end{table}

\subsection{Ablation Study} \label{subsec:ablation}
We conduct the ablation study by using Conditional DETR~\cite{meng2021conditional} as the baseline. The CNN backbone is ResNet-50~\cite{he2016deep}, and the training epoch nubmer is $12$. The performances are evaluated on COCO {\em val2017}~\cite{lin2014coco}. We mainly study the effects of the key design: group-wise one-to-many assignment, separate self-attention, and group number.

\paragraph{Group-wise one-to-many assignment and separate self-attention.} Table~\ref{tab:o2mimplementations1250} shows how group-wise one-to-many (o2m) assignment and separate self-attention make contributions. In comparison to the baseline (a), group-wise o2m assignment improves the mAP score from $32.6$ mAP to $34.8$ mAP: with the gain $2.2$. The separate self attention (Sep. SA) further gets a $2.8$ mAP gain. In addition, we report naive one-to-many assignment. The results are very poor, which is reasonable in that there are duplicate predictions and there is a lack of scoring mechanisms for demoting them. The results suggest that both group-wise o2m assignment and separate self-attention are effective.

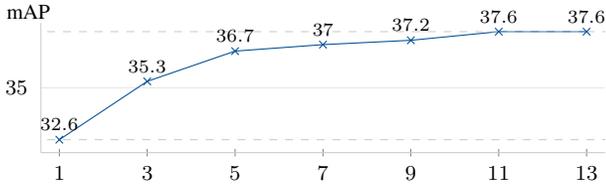
\begin{figure}[t]
\centering
\begin{tikzpicture}[font=\footnotesize]
\pgfplotsset{set layers, compat=1.8, 
every axis/.append style={
font=\scriptsize,
}
}
    \begin{groupplot}[
        group style={
            group size=1 by 2,
            vertical sep=0pt
        },
        axis lines = left,
        every outer y axis line/.style={draw=gray!40},
        every outer x axis line/.style={draw=gray!40},
        yticklabels={},
    	footnotesize,
        scale only axis,
    	x post scale=1.5,
        xticklabels={$1$, $3$, $5$, $7$, $9$, $11$, $13$, $15$},
    	xmin=1,
    	xmax=13,
    ylabel={mAP},
    nodes near coords, 
	nodes near coords align={vertical},
    	scaled ticks=false,
    ]
        \nextgroupplot[
          y post scale=0.4,
          xtick=data,
          ymin=32,
          ymax=38,
          ytick style={draw=none},
          enlargelimits=false,
    	  enlarge x limits=0.035,
    	  extra y ticks={35},
    	  extra y tick labels={$35$},
    	  extra y tick style={grid=major, draw=gray!40},
          grid style={line width=.1pt, draw=gray!20},
          ylabel style = {rotate=270, font=\footnotesize, anchor=north, yshift=35pt, xshift=13.5pt}
        ]
    
        \addplot[draw=mediumelectricblue, mark=x] 
	coordinates{
	(1, 32.6)
	(3, 35.3)
	(5, 36.7)
	(7, 37.0)
	(9, 37.2)
	(11, 37.6)
	(13, 37.6)
	};
        \draw[gray!40, dashed] (axis cs:0.,32.6) -- (axis cs:14,32.6);
        \draw[gray!40, dashed] (axis cs:0.,37.6) -- (axis cs:14,37.6);
    \end{groupplot}
\end{tikzpicture}
\caption{\textbf{Influence of group number.} The $x$-axis is the number of groups. It can be seen that the performance becomes stable when the number of groups reaches $11$.} 
\label{fig:num_groups}
\end{figure}

\begin{table}[t]
  \centering
    \setlength{\tabcolsep}{8pt}
    \renewcommand{\arraystretch}{1.25}
    \footnotesize
      \caption{\textbf{Effects of group-wise one-to-many assignment and separate self-attention.} (a) baseline: one-to-one assignment with $300$ object queries. (b) naive one-to-many assignment with $3300$ object queries for training and inference. (c) group-wise one-to-many assignment and no separate self-attention with $11$ groups of $300$ queries, inference with a group of $300$ queries. (d) group-wise one-to-many assignment and separate self-attention with $11$ groups of $300$ queries, inference with a group of $300$ queries. o2m = one-to-many, Sep. SA = separate self-attention.}
  \label{tab:o2mimplementations1250}
  \vspace{2mm}
  \begin{tabular}{ccc|cccc}
    \shline
    & o2m & Sep. SA & mAP & AP$_{s}$ & AP$_{m}$ & AP$_{l}$ \\
    \hline
   (a) & $\times$ & $\times$ & $32.6$ & $14.4$ & $34.9$ & $48.6$ \\
    \hline
   (b) & Naive & $\times$ & $8.4$ & $8.0$ & $13.2$ & $13.3$ \\
   (c) & Group & $\times$ & $34.8$ & $16.4$ & $37.7$ & $51.4$ \\
    (d) &Group  & $\checkmark$ & $\mathbf{37.6}$ & $\mathbf{18.2}$ & $\mathbf{40.7}$ & $\mathbf{55.9}$ \\
    \shline
  \end{tabular}
  \vspace{-1mm}
\end{table}
\vspace{1mm}
\noindent\textbf{Group number.} Figure~\ref{fig:num_groups} shows the influence of the number of groups $K$ in Group DETR. The detection performance improves when increasing the number of groups, and becomes stable when the group number reaches $11$. Thus, we adopt $K=11$ by default in Group DETR in our experiments.

\section{Related Works}
There are two main lines for accelerating DETR training: modify {\em cross-attention} and stabilize {\em one-to-one assignment}. The two are complementary and can be combined to further boost the performance.

\vspace{1mm}
\noindent\textbf{Modifying cross-attention.} Cross-attention module aims to collect the information from the image features useful to classification and localization. Various methods are proposed to select the informative image regions more efficiently and effectively~\cite{gao2021fast,chen2022conditional,wang2021anchor,yao2021efficient,liu2022dab,gao2022adamixer}. For example, Deformable attention~\cite{deformable-detr} selects the highly informative positions dynamically according to the previous decoder embedding. Conditional DETR~\cite{chen2022conditional} instead continues to use the normal global attention, and dynamically computes the spatial attention to softly select the informative regions. SMCA~\cite{gao2021fast} uses the Gaussian-like weight for spatial modulation.

\vspace{1mm}
\noindent\textbf{Stabilizing one-to-one assignment.} DETR~\cite{detr} relies on one-to-one assignment, where each ground-truth object is assigned to a single prediction through building a bipartite matching between the predictions and the ground-truth objects. DN-DETR~\cite{li2022dn} finds the assignment process is unstable and attributes the slow convergence issue to the instabilities. Thus, DN-DETR~\cite{li2022dn} introduces groups of noisy queries by adding noises to ground-truth objects, to stabilize the assignment, leading to faster convergence. DINO~\cite{zhang2022dino} makes further improvement through contrastive denoising training to generate both positive and negative noise queries with different noise levels. Our approach studies the assignment mechanism instead for introducing more supervision.

\paragraph{One-to-many assignment.} One-to-many assignment is widely adopted in deep detectors~\cite{ren2015faster,he2017mask,lin2017focal,tian2019fcos}, and has attracted a lot of interest~\cite{zhang2020bridging,kim2020probabilistic,zhu2020autoassign,ge2021ota,chen2021you,wang2021end,sun2021makes}. For example, Faster R-CNN~\cite{ren2015faster} and FCOS~\cite{tian2019fcos} produce multiple positive anchors and pixels for each ground-truth object. In this paper, we investigate one-to-many assignment in a feasible manner for the end-to-end detector DETR. 

Concurrent with our work, $\mathcal{H}$-DETR~\cite{jia2022detrs} also uses one-to-many assignment to speed up DETR training convergence. Our Group DETR and $\mathcal{H}$-DETR are related, but different: (1) Group DETR introduces group-wise one-to-many assignment with separate self-attention with the same number of object queries in each group. $\mathcal{H}$-DETR adopts hybrid assignments in two different groups: One group uses one-to-one assignment and another uses one-to-many assignment with more object queries. (2) All the decoders in Group DETR can be used for inference. But the additional decoder in $\mathcal{H}$-DETR is not directly used and requires NMS for inference. (3) During training, our architecture introduces one parameter: the number of groups. In contrast, $\mathcal{H}$-DETR introduces the number of additional queries and the number of additional positive queries.

DETA~\cite{ouyang2022nms} is another concurrent work with our Group DETR. DETA directly uses one-to-many assignment and brings NMS back to DETR frameworks. While our method provides group-wise one-to-many assignment and maintains end-to-end detection.

\section{Conclusion}
The key points in Group DETR include group-wise one-to-many assignment and parallel self-attention. The success stems from involving more groups of object queries as an addition to the primary group of object queries, and thus introducing more supervision. Group-wise assignment mechanism makes sure that the competition among predictions happens within each group separately, and separate self-attention eases the training, Thus, the NMS pose-processing is not necessary, and the inference process is kept the same as normally trained DETR and not dependent on the group design. Our approach is simple, easily implemented, and general.

\noindent \textbf{Acknowledgements.} 
This work is supported by the Sichuan Science and Technology Program (2023YFSY0008), National Natural Science Foundation of China (61632003, 61375022, 61403005), Grant SCITLAB-20017 of Intelligent Terminal Key Laboratory of SiChuan Province, Beijing Advanced Innovation Center for Intelligent Robots and Systems (2018IRS11), and PEK-SenseTime Joint Laboratory of Machine Vision.

\section*{Appendix}
\setcounter{section}{0}
\renewcommand{\thesection}{\Alph{section}}
\section{More Details and Results}
\subsection{Datasets and Evaluation Metrics}
We perform the object detection and instance segmentation experiments on the COCO $2017$~\cite{lin2014coco} dataset, which contains about $118$K training (\textit{train2017}) images, $5$K validation (\textit{val2017}) images, and $20$K testing (\textit{test-dev}) images. Following the common practice, we train our model on COCO \textit{train2017} and report the standard mean average precision (mAP) result (box mAP for object detection and mask mAP for instance segmentation) on the COCO \textit{val2017} dataset under different IoU thresholds (from $0.5$ to $0.95$) and object scales (small, medium, and large). We also report the result on COCO \textit{test-dev} with a large foundation model (ViT-Huge~\cite{zhai2022scaling,He_2022_CVPR,chen2022cae}).

We perform multi-view 3D object detection experiments on the nuScenes~\cite{caesar2020nuscenes} dataset, which contains $1000$ driving sequences. There are $700$ for \textit{train} set, $150$ for \textit{val} set and $150$ for \textit{test} set. We report the standard nuScenes Detection Score (NDS) and mean Average Precision (mAP) result on the nuScenes \textit{val} set.

\begin{table*}
\centering
  \setlength{\tabcolsep}{5.5pt}
    \renewcommand{\arraystretch}{1.25}
    \footnotesize
  \caption{\textbf{  Our method achieves $64.5$ mAP on the COCO {\em test-dev}.}}
  \vspace{1mm}
  \begin{tabular}{l|cccc|c}
    \shline
    Method & {\color{gray} \#Params} & {\color{gray} Encoder Pretraining Data} & {\color{gray} Detector Pretraining Data} &  {\color{gray} w/ Mask} & mAP \\
    \hline
    Swin-L (HTC++)~\cite{liu2021swin}  & {\color{gray} $284$M} & {\color{gray} IN-$22$K ($14$M)} & {\color{gray} n/a} & {\color{gray} \checkmark} & $58.7$ \\
    DyHead (Swin-L)~\cite{dai2021dynamic}  & {\color{gray} $213$M} & {\color{gray} IN-$22$K ($14$M)} & {\color{gray} n/a} & {\color{gray} \checkmark} & $60.6$ \\
    Soft-Teacher (Swin-L)~\cite{xu2021end} & {\color{gray} $284$M} & {\color{gray} IN-$22$K ($14$M)} & {\color{gray} COCO-unlabeled + O365} & {\color{gray} \checkmark} & $61.3$ \\
    GLIP (DyHead)~\cite{li2022grounded}  & {\color{gray} $\geq$$284$M} & {\color{gray} IN-$22$K ($14$M)} & {\color{gray} FourODs + GoldG + Cap24M} & {\color{gray} $\times$} & $61.5$ \\
    Florence (CoSwin-H)~\cite{zhang2022glipv2} & {\color{gray} $\geq$$637$M} & {\color{gray} FLD-$900$M (900M)} & {\color{gray} FLD-$9$M} & {\color{gray} $\times$} & $62.4$ \\
    GLIPv2 (CoSwin-H)~\cite{zhang2022glipv2} & {\color{gray} $\geq$$637$M} & {\color{gray} FLD-$900$M (900M)} & {\color{gray} merged data$^b$} & {\color{gray} $\checkmark$} & $62.4$ \\
    SwinV2-G (HTC++)~\cite{liu2022swin} & {\color{gray} $3.0$B} & {\color{gray} IN-$22$K + ext-$70$M ($84$M)} & {\color{gray} O365} & {\color{gray} \checkmark} & $63.1$ \\
    DINO-{\em $5$scale} (Swin-L)~\cite{zhang2022dino} & {\color{gray} $218$M} & {\color{gray} IN-$22$K ($14$M)} & {\color{gray} O365} & {\color{gray} $\times$} & $63.3$ \\
    BEIT-3 (ViTDet)~\cite{wang2022image} & {\color{gray} $1.9$B} & {\color{gray} merged data$^a$} & {\color{gray} O365} & {\color{gray} \checkmark} & $63.7$ \\
    FD-SwinV2-G (HTC++)~\cite{wei2022contrastive} & {\color{gray} $3.0$B} & {\color{gray} IN-$22$K + IN-$1$K + ext-$70$M ($85$M)} & {\color{gray} O365} & {\color{gray} \checkmark} & $64.2$ \\
    FocalNet-H (DINO-{\em $5$scale})~\cite{yang2022focal} & {\color{gray} $746$M} & {\color{gray} IN-$22$K ($14$M)} & {\color{gray} O365} & {\color{gray} $\times$} & $64.3$ \\
    $Co$-Deformable-DETR (MixMIM-g)~\cite{liu2022mixmim,zong2022detrs} & {\color{gray} $1.0$B} & {\color{gray} IN-$1$K ($1$M)} & {\color{gray} O365} & {\color{gray} $\times$} & $64.5$ \\
    EVA (CMask R-CNN)~\cite{fang2022eva,cai2019cascade,he2017mask} & {\color{gray} $\geq$$1.0$B} & {\color{gray} merged-$30$M$^c$} & {\color{gray} O365} & {\color{gray} \checkmark} & $64.7$ \\
    InternImage-H (DINO-{\em $5$scale})~\cite{wang2022internimage,su2022towards,zhang2022dino} & {\color{gray} $2.18$B} & {\color{gray} merged data$^d$} & {\color{gray} O365} & {\color{gray} $\times$} & $65.4$ \\
    \hline
    \textbf{ViT-Huge + Group DETR (DINO-{\em $4$scale})} & {\color{gray} $629$M} & 
    {\color{gray} IN-$1$K ($1$M)} & {\color{gray} O365} & {\color{gray} $\times$} & $64.5$ \\
    \shline
  \end{tabular}
  \vspace{1mm}
  \label{tab:leaderboard}
  \\All the results are achieved with test time augmentation. In the table, we follow the notations for various datasets used in DINO~\cite{zhang2022dino} and FocalNet~\cite{yang2022focal}. `w/ Mask' means using mask annotations when finetuning the detectors on COCO~\cite{lin2014coco}. And for the baseline DINO, we adopt the {\em $4$scale} version~\cite{zhang2022dino}.\\
  `merged data$^a$': IN-$22$K + Image-Text ($35$M) + Text ($160$GB). `merged data$^b$': FourODs + INBoxes + GoldG + CC15M + SBU. \\
  `merged-$30$M$^c$': IN-$21$K + O365 + COCO + ADE20K + CC15M. `merged data$^d$': Laion-400M + YFCC-15M + CC12M.
  \end{table*}

\subsection{Implementation Details}
Our Group DETR adopts multiple groups of object queries. Each group shares the same architectures and numbers of object queries\footnote{When applying Group DETR to DN-DETR~\cite{li2022dn} and DINO~\cite{zhang2022dino}, we add the corresponding query denoising task in each group to keep the same architecture with the original implementation.}. It resembles data augmentation with automatically-learned object query augmentation and is also equivalent to simultaneously training parameter-sharing networks of the same architecture.

In one-stage DETR frameworks, including Conditional DETR~\cite{meng2021conditional}, DAB-DETR~\cite{liu2022dab}, DN-DETR~\cite{li2022dn}, and DAB-Deformable-DETR~\cite{liu2022dab,deformable-detr}, we can easily implement Group DETR by adopting multiple groups of learnable object queries. While the situation is different in two-stage DETR frameworks, such as DINO~\cite{zhang2022dino}. The initializations of object queries are dependent on the top-$N$ predicted boxes of the first stage. To make the object queries in multiple groups similar to each other, we construct multiple pairs of classification and regression prediction heads in the first stage, each pair of which provides initialization for the object queries in the corresponding group. As for model inference, we only need one pair of these prediction heads, the same as the original model.

\subsection{More Results of DN-DETR}
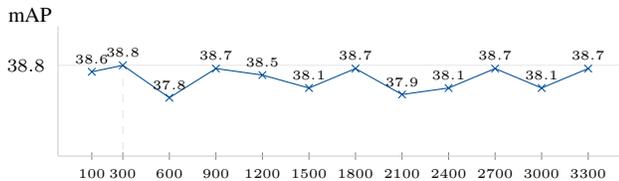
\begin{figure}[t]
\centering
\begin{tikzpicture}[font=\scriptsize]
\pgfplotsset{set layers, compat=1.8, 
every axis/.append style={
font=\scriptsize,
}
}
    \begin{groupplot}[
        group style={
            group size=1 by 2,
            vertical sep=0pt
        },
        axis lines = left,
        every outer y axis line/.style={draw=gray!40},
        every outer x axis line/.style={draw=gray!40},
        yticklabels={},
    	footnotesize,
        scale only axis,
    	x post scale=1.5,
        xticklabels={$100$, $300$, $600$, $900$, $1200$, $1500$, $1800$, $2100$, $2400$, $2700$, $3000$, $3300$},
    	xmin=1,
    	xmax=3400,
    ylabel={mAP},
    nodes near coords,
every node near coord/.append style={xshift=0pt,yshift=10pt,anchor=north,font=\tiny},
every x tick label/.append style={font=\tiny},
every y tick label/.append style={font=\scriptsize},
    	scaled ticks=false,
    ]
        \nextgroupplot[
          y post scale=0.4,
          xtick=data,
          ymin=36,
          ymax=40,
          ytick style={draw=none},
          enlargelimits=false,
    	  enlarge x limits=0.035,
    	  extra y ticks={38.8},
    	  extra y tick labels={$38.8$},
    	  extra y tick style={grid=major, draw=gray!40},
          grid style={line width=.1pt, draw=gray!20},
          ylabel style = {rotate=270, font=\footnotesize, anchor=north, yshift=35pt, xshift=13.5pt}
        ]
        \addplot[draw=mediumelectricblue, mark=x] 
	coordinates{
	(100, 38.6)
	(300, 38.8)
	(600, 37.8)
        (900, 38.7)
        (1200, 38.5)
        (1500, 38.1)
        (1800, 38.7)
        (2100, 37.9)
        (2400, 38.1)
        (2700, 38.7)
        (3000, 38.1)
        (3300, 38.7)
	};
        \draw[gray!20, dashed] (axis cs:300,0) -- (axis cs:300,38.8);
    \end{groupplot}
\end{tikzpicture}
  \caption{\textbf{Results of DN-DETR with different number of denoising queries.} We show the detection performances (mAP) on MS COCO~\cite{lin2014coco} of adopting different number of denoising queries in DN-DETR.}
  \label{figure:numberdnqueries}
\end{figure}

\paragraph{Results of DN-DETR with different numbers of denoising queries.} We conduct experiments with different numbers of denoising queries in DN-DETR~\cite{li2022dn}. The results in Figure~\ref{figure:numberdnqueries} suggest that increasing the number of denoising queries can not achieve further improvements and show unstable performances. The effects of denoising queries differ from the ones of Group DETR (Figure~\ref{fig:comparewithdndetr} in the main paper). We choose to use $100$ denoising queries in our experiments in Table~\ref{tab:main1} and Table~\ref{tab:main2} in the main paper by following the setting in the original paper~\cite{li2022dn}. To make direct comparisons with DN-DETR~\cite{li2022dn}, we report the best results across different numbers of denoising queries in Figure~\ref{figure:numberdnqueries} ($38.8$ mAP). 

\subsection{Applying Group DETR to SAM-DETR series}
We also apply Group DETR to another stream of work to accelerate DETR training, SAM-DETR~\cite{zhang2022accelerating} and SAM-DETR++~\cite{zhang2022semantic}. The results are given in Table~\ref{tab:samdetrwithgroupdetr}. Improvements on SAM-DETR~\cite{zhang2022accelerating} (gains: $3.1$ mAP with 12e and $1.9$ mAP with 50e) and SAM-DETR++~\cite{zhang2022semantic} (gains: $2.2$ mAP with 12e and $1.3$ mAP with 50e) show that Group DETR is complementary to them as well.

\section{More Comparisons on COCO \textit{test-dev}}
\paragraph{Settings.} To compare state-of-the-art results on COCO \textit{test-dev}, we follow DINO~\cite{zhang2022dino} to build our model with a large foundation model, ViT-Huge. We follow its training pipeline and settings: 
(i) pre-train~\cite{chen2022cae} and fine-tune the ViT-Huge on ImageNet-1K~\cite{deng2009imagenet}, 
(ii) pre-train the whole detector on Object365~\cite{shao2019objects365} for $24$ epochs with $64$ A100 GPUs, 
and (iii) finetune the detector on COCO~\cite{lin2014coco} for $20$ epochs with $32$ A100 GPUs. When pre-training the detector on Object365, 
we follow DINO~\cite{zhang2022dino} to only leave the first $5$k out of $80$k validation images as the validation set 
and add the other images to the training set. 
We also use other schemes when training the detector on Object365 and COCO, 
such as enlarging the image size to $1.5\times$ when finetuning and adopting test time augmentation. 
In addition, we apply the exponential moving average (EMA) technique~\cite{tarvainen2017mean}, use CDN queries~\cite{zhang2022dino}, and adopt $11$ groups with Group DETR during detector pre-training and fine-tuning. When fine-tuning the detector on COCO, we find that applying learning rate decay~\cite{clark2020electra,bao2021beit,He_2022_CVPR,chen2022cae} for the components of the detector gives a $\sim$$0.9$ mAP gain on COCO. During testing, we adopt test time augmentation with various scales and their flipped counterparts and perform fusion\footnote{According to our experiments, the fusion on the query features builds a robust feature across different scales and gives a $\sim$$0.8$ mAP improvement.} on the query features and the final predictions~\cite{zhang2022dino}.

\begin{table}
  \centering
    \setlength{\tabcolsep}{2pt}
    \renewcommand{\arraystretch}{1.5}
    \footnotesize
  \caption{\textbf{Effectiveness of Group DETR on SAM-DETR and SAM-DETR++.} All experiments adopt ResNet-50~\cite{he2016deep} and evaluate on COCO {\em val2017}~\cite{lin2014coco}.}
  \vspace{2mm}
  \begin{tabular}{l|c|ccc}
    \shline
    Model     & w/ Group & Epochs & mAP &\\
    \hline
    \multirow{2}{*}{SAM-DETR} &  & 12 & $33.1$  &\\
     & \checkmark & 12 & \reshl{$36.2$}{+}{$\mathbf{3.1}$} & \\
     \hline
     \multirow{2}{*}{SAM-DETR} &  & 50 & $39.8$ &\\
     & \checkmark & 50 & \reshl{$41.7$}{+}{$\mathbf{1.9}$} &\\
     \hline
    \multirow{2}{*}{SAM-DETR++} &  & 12 & $41.1$ &\\
     & \checkmark & 12 & \reshl{$43.3$}{+}{$\mathbf{2.2}$} &\\
     \hline
     \multirow{2}{*}{SAM-DETR++} &  & 50 & $46.1$ &\\
     & \checkmark & 50 & \reshl{$47.4$}{+}{$\mathbf{1.3}$} &\\
    \shline
  \end{tabular}
  \label{tab:samdetrwithgroupdetr}
  \vspace{-2mm}
\end{table}

\paragraph{Results.} Table~\ref{tab:leaderboard} shows the results. Our model is the first to achieve $\mathbf{64.5}$ mAP on COCO {\em test-dev}. Only pre-training the ViT-Huge on ImageNet-1K~\cite{deng2009imagenet}, our model can outperform other methods with larger models (e.g., BEIT-3~\cite{wang2022image} and SwinV2-G~\cite{liu2022swin,wei2022contrastive}) and more pre-training data. Models such as EVA~\cite{fang2022eva} and InterImage-H~\cite{wang2022internimage}, with larger foundation models (ViT-giant~\cite{zhai2022scaling} or InterImage-H~\cite{wang2022internimage}) and more data~\cite{deng2009imagenet,changpinyo2021conceptual,sharma2018conceptual,zhou2017ade20k,thomee2016yfcc100m,schuhmann2021laion}, give higher results ($64.7$ mAP and $65.4$ mAP) than our model. We expect that our results will be further improved with more pre-training data and larger models.

{\small
\bibliographystyle{ieee_fullname}
\bibliography{egbib}
}

\end{document}